\documentclass{itatnew}
\usepackage{mathtools}
\usepackage{enumitem}
\usepackage[noabbrev,capitalize]{cleveref}
\usepackage{xcolor}
\usepackage{booktabs}

\usepackage{graphicx}
\usepackage{subcaption} 

\def\XXX#1{{\textcolor{red}{XXX #1}}}
\def\hideXXX#1{}

\def\furl#1{\footnote{\url{#1}}}
\def\hidelong#1{#1}

\def\perscite#1{\cite{#1}}
\def\inparcite#1{\cite{#1}}

\title{
Presenting Simultaneous Translation in Limited Space}

\author{Dominik Mach{\' a}{\v c}ek, Ond{\v r}ej Bojar}
\institute{Charles University \\
Faculty of Mathematics and Physics \\
Institute of Formal and Applied Linguistics \\
\email{\{machacek,bojar\}@ufal.mff.cuni.cz}
}

\date{}

\begin{document}
\maketitle
\begin{abstract}
Some methods of automatic simultaneous translation of a long-form speech allow revisions of outputs, trading accuracy for low latency. Deploying these systems for users faces the problem of presenting subtitles in a limited space, such as two lines on a television screen. The subtitles must be shown promptly, incrementally, and with adequate time for reading. We provide an algorithm for subtitling. Furthermore, we propose a way how to estimate the overall usability of the combination of automatic translation and subtitling by measuring the quality, latency, and stability on a test set, and propose an improved measure for translation latency. 
\end{abstract}

\section{Introduction}

The quality of automatic speech recognition and machine translation of texts is constantly increasing. It leads to an opportunity to connect these two components and use them for spoken language translation (SLT).
The output of the SLT system can be delivered to users either as speech or text. In simultaneous SLT, where the output has to be delivered during the speech with as low delay as possible, there is a trade-off between latency and quality. With textual output, it is possible to present users with early, partial translation hypotheses in low latency, and correct them later by final, more accurate updates, after the system receives more context for disambiguation, or after a secondary big model produces its translation. Rewriting brings another challenge, the stability of output. If the updates are too frequent, the user is unable to read the text. The problem of unstable output could be solved by using big space for showing subtitles. The unstable, flickering output would appear only at the end, allowing the user to easily ignore the flickering part and read only the stabilized part of the output. However, in many situations, the space for subtitles is restricted. For example, if the users have to follow the speaker and slides at the same time, they lack mental capacity for searching for the stabilized part of translations. It is, therefore, necessary to put the subtitles and slides on the same screen, restricting the subtitling area to a small window.


In this paper, we propose an algorithm for presenting SLT subtitles in limited space, a way for estimating the overall usability of simultaneous SLT subtitling in a limited area, and an improved translation latency measure for SLT comparison. 
\hidelong{\cref{pipeline} describes the properties of SLT for use with our subtitler. \cref{subtitler} details the main new component for presenting a text stream as readable subtitles. \cref{empirical} proposes the estimation of the usability of the subtitling of multiple realistic SLT systems.
We conclude the paper in \cref{conclusion}.
}

\section{Re-Translating Spoken Language Translation}
\label{pipeline}



%


Our subtitler solves the problem of presentation of SLT output with a re-translating early hypothesis, similarly to \perscite{niehues2016inter,arivazhagan2019retranslation,Arivazhagan2020RetranslationVS}. Although it can also present the subtitles from the automatic speech recognition (ASR) that re-estimates the early hypothesis, or generally any audio-to-text processor, we limit ourselves only SLT in this paper for brevity.


\hidelong{\subsection{Stable and Unstable Segments}}

\hidelong{
SLT systems output a potentially infinite stream of segments containing the beginning and final timestamps of an interval from the source audio, and the translated text in the interval. We assume that the segments can be marked as \emph{stable} and \emph{unstable},
depending on whether the system has the possibility to change them or not. 
This is a realistic assumption because the ASR and SLT systems usually process a limited window of the source audio. Whenever a part of source audio exceeds this window, the corresponding output becomes stable.
}

\begin{figure*}[t]
\centering
\includegraphics[width=\textwidth]{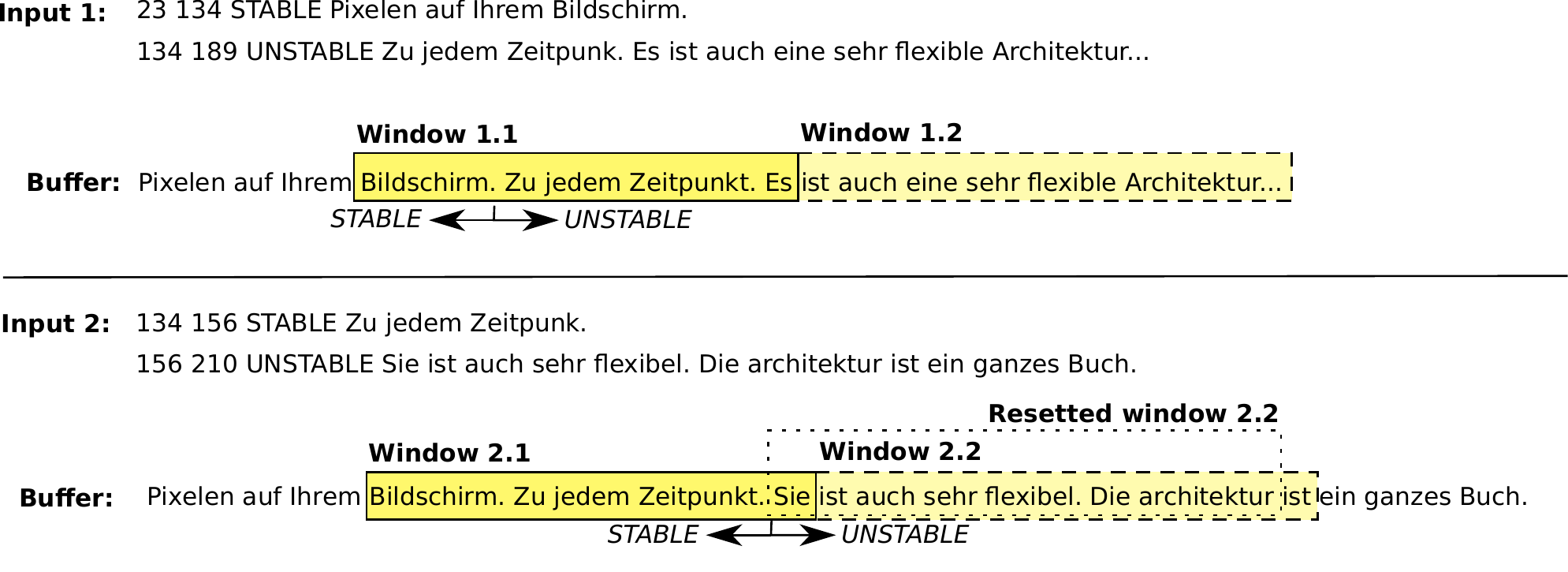}
\caption{Illustration of speech translation subtitling in two subsequent inputs from SLT. The input arrives as a sequence of quadruples: segment beginning time, segment end time, stable/unstable flag, text. The rectangles indicate the content of the subtitling area of one line.}
\label{fig:illustration}
\end{figure*}

\begin{figure}
    \centering
\includegraphics[width=0.4\textwidth]{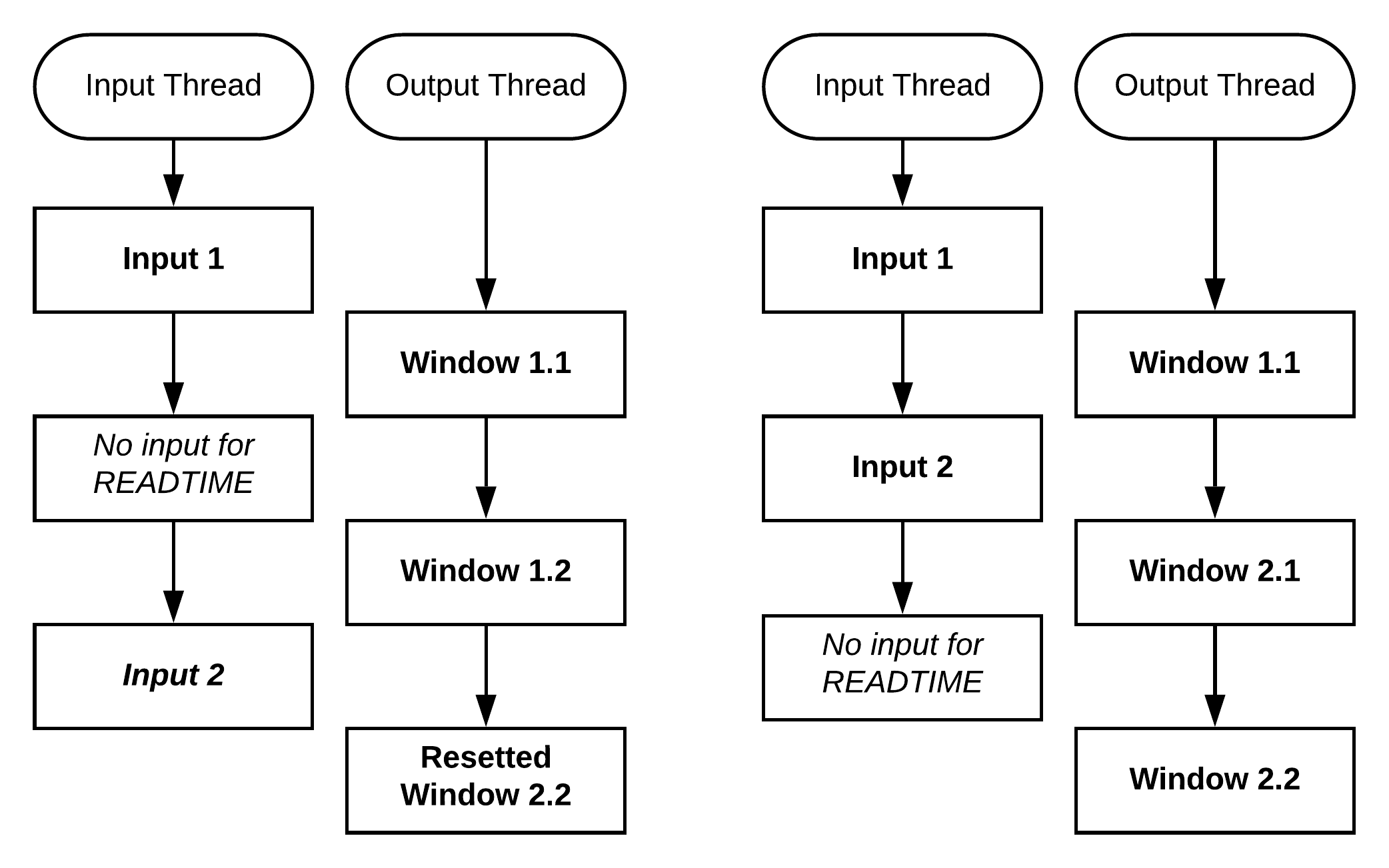}
    \caption{Subtitler processing of the inputs in \cref{fig:illustration} with different timings. In the left one, Input 2 changes the word ``Es'', which has been read by the user and scrolled away and causes a reset of a window start. In the right one, the word ``Es'' is changed in the window on the current display.}
    \label{fig:diagrams}
\end{figure}

\section{Subtitler}
\label{subtitler}

This section presents the design and algorithm of ``subtitler''. 

The subtitler is a cache on a stream of input messages aiming
to satisfy the following conflicting needs:
\begin{itemize}[nosep]
\item The output should be presented with the lowest possible delay to
achieve the effect of simultaneous translation as much as possible.
\item The flickering of the partial outputs is partially desired because it
highlights the simultaneity of the translation and comforts the user in knowing
that the system is not stuck.
\item The flickering should be minimized. If some output was presented at a
position of the screen, it should keep the position until it is outdated.
\item The user must have enough time to read the message.
\item Only a predefined space of $w$ (width) characters and $h$ (height) lines
are available.
\end{itemize}

Given an input stream of stable and unstable segments as described
above, the subtitler emits a stream of ``subtitle windows''. On every update, the former window is replaced by a new one. 


The basic operation of subtitler is depicted in \cref{fig:illustration,fig:diagrams}. The elements of subtitler are a
buffer of input segments,  a presentation window, and two independent
processing threads. 


The buffer is an ordered list of segments. The
presentation window is organized as a list of text lines of the required width and
count. The count corresponds to the height of subtitling window plus one, to allow scrolling-up the top line after displaying it for minimum reading time. This line view is regenerated whenever needed from the current starting position of
the window in the buffer, wrapping words into lines. 

The \emph{input thread} receives the input stream and updates the buffer.
It replaces outdated segments with their new versions, extends the buffer, and
removes old unnecessary segments. If an update happens within
or before the current position of the presentation window, the \emph{output
thread} is notified for a forced update.

Independently, the \emph{output thread} updates the position
of the presentation window in the buffer, obeying the following timeouts and triggers: 
\begin{itemize}[nosep]
\item On forced updates, the output thread detects if any content changed 
before the beginning of the already presented window, which would cause
a \emph{reset}.
In that case, the window position on the window buffer has to be
moved back, and the content for the user can no longer be presented
incrementally. Instead, the beginning of the first line in the window shows
a newer version of an old sentence that has already scrolled away. 
\item If the first line of the presentation window has not been changed for a
minimum reading time and if there is any input to present in the extra line of the
window, the window is ``scrolled'' by one line, i.e., the first line is discarded,
the window starting position within the buffer is updated, and the extra line is shown as the last line of the window.
\item If the whole presentation window has not been changed for a long time, e.g., 5 or 20 seconds, it is blanked by emitting empty lines.
\end{itemize}

\subsection{Timing Parameters}

The subtitler requires two timing parameters. A line of subtitles is displayed to a user for a ``minimum reading time'' before it can be scrolled away. If no input arrives for a ``blank time'', the subtitling window blanks to indicate it and to prevent the user from reading the last message unnecessarily. 

We suggest adopting the minimum reading time parameter from the standards for subtitling films and videos (e.g., \inparcite{fotios}), before standards for simultaneous SLT subtitling will be established.
\perscite{Szarkowska2018ViewersCK} claim that 15 characters per second is a standard reading time in English interlingual subtitling of films and videos for deaf and hard hearing. The standards in other European regions are close to 15 characters per second. We use this value for the evaluation in this work. 



\section{Estimating Usability}
\label{empirical}

The challenges in simultaneous SLT are quality, latency, and stability \cite{niehues2016inter,arivazhagan2019retranslation}. All of these properties are critical for the overall usability of the SLT system. The quality of translation is a property of the SLT system. The subtitler has no impact on it.
The minimum reading time ensures the minimum level of stability, ensuring that every stable content is readable, and may increase the latency if the original speech is faster than reading time.
The size of the subtitling window and timing parameters affect overall latency and stability.
The bigger the window, the longer updates of translations fit into it without a reset. The \hideXXX{myslim, ze tu ma byt minimum reading time, vsude, az do konce odstavce: DM: nema. Blanking time ovlivnuje resety. Mozna ma byt jak minimum reading, tak blanking... OB: tak tedy oba, protoze minimum reading urcite. Kdyz uz radku prectes (tj. uplyne minumum reading), tak se odroluje  a hrozi reset. Blanking typicky *nezazijes*, protoze je az po 20 vterinach bez vstupu.
OB: napisme proste "timing constants"? timing parameters OK, 
DM: Ale nekdy jo. Spocital jsem resety pro 3 radky a pro ASR, kde se vsechny maji vejit, to nekde udelalo reset kvuli tomu. OK, chapu. Myslim, ze ted je to dobre. Pozor, dal jsou dalsi veci.}
timing parameters determine how long the content stays unchanged in the window before scrolling. A small subtitling window or a short reading or blanking time may cause a reset. Every reset increases latency because it returns to the already displayed content. On the other hand, the significant latency may improve stability by skipping the early unstable hypotheses and present only the stable ones.


We provide three automatic measures for assessing the practical usability of simultaneous SLT subtitling on the test set. The automatic evaluation may serve for a rough estimation of the usefulness, or for selection of the best candidate setups. We do not provide a strict way to judge which SLT system and subtitling setup are useful and which are not. The final decision should ideally consider the particular display conditions, expectations, and needs of the users, and should be based on a significant human evaluation.

\subsection{Evaluation Measures}

For quality, we report an automatic machine translation measure BLEU computed by sacrebleu \cite{post-2018-call} after automatic sentence alignment using mwerSegmenter \cite{segmentation:matusov:2005:iwslt}. BLEU is considered as correlating with human quality judgement. The higher BLEU, the higher translation quality.

To explain the measure of latency and stability, let us use the terminology of \perscite{arivazhagan2019retranslation}. The \emph{EventLog} is an ordered list of \emph{events}. The $i^{th}$ \emph{event} is a triple $s_i, o_i, t_i$, where $s_i$ is the source text recognized so far, $o_i$ is the current SLT output, and $t_i$ is the time when this event was produced. Source and output, $s_i$ and $o_i$, are sequences of tokens. Let us denote $c(o_i)$ a transformation of a token sequence into a sequence of characters, including spaces and punctuation. Let $I$ be the number of all events, with an update either in source or output, and $T$ the number of events with an update in translation.

\subsubsection{Character Erasure}

To evaluate how many updates fit into the subtitling window, we define \emph{character erasure} (cE). It is the number of characters that must be deleted from the tail of the current translation hypothesis to update it to a new one. If a new translation only appends words to the end, the erasure is zero. The \emph{character erasure} is  
cE$(i) = |c(o_{i-1})| - |LCP(c(o_i),c(o_{i-1}))|$, where the $LCP$ stands for the longest common prefix. The average character erasure is AcE = $1/T \sum_{i=1}^I $cE$(i)$. It is inspired by the normalized erasure (NE) by \perscite{arivazhagan2019retranslation}, but we do not divide it by the output length in the final event, but only by the number of translation events.

\subsubsection{Translation Latency with Sentence-Alignment Catch-up}

The translation latency may be measured with the use of a \emph{finalization event} of the $j$-th word in output. It is $f(o,j) = min_i$ such that $o_{i',j'} = o_{I,j'}$ $\forall i'\ge i$ and $\forall j'\le j$. In other words, the word $j$ is finalized in the first event $i$, for which the word $j$ and all the preceding words $j'$ remain unchanged in all subsequent events $i'$.

The translation latency of output word $j$ is the time difference of the finalization event of the word $j$ in the output and its \emph{corresponding} word $j^*$ in the source. \perscite{arivazhagan2019retranslation} estimate the source word simply as $j^* = (j/|o_I|) |s_I|$. This is problematic if the output is substantially shorter than input, because then it may incorrectly base the latency on a word which has not been uttered yet, leading to a negative time difference. 
A proper word alignment would provide the most reliable correspondence. However, we propose a simpler and appropriately reliable solution. The following improved measure is our novel contribution. We use it to compare the SLT systems.

We utilize the fact that our ASR produces punctuated text, where the sentence boundaries can be detected. The sentences coming from SLT and ASR in their sequential order are parallel. They can be simply aligned because our SLT systems translate the individual sentences and keep the sentence boundaries.
If the SLT does not produce individual sentences, then we use a rule-based sentence segmenter, e.g. from \perscite{moses}, and must be aware of the potential inaccuracy.

We use the sentence alignment for a catch-up, and the simple temporal correspondence of \perscite{arivazhagan2019retranslation} only within the last sentence. 
To express it formally, let us assume that the EventLog has also a function $S(o,j)$, returning the index of the sentence containing the word $j$ in $o$, and $L(o,k)$, the length of the sentence $k$ in $o$. Let $x(j) = j-\sum_{i=1}^{S(o,j)-1} L(o,i)$ be the index of an output word $j$ in its sentence. Then we define our caught-up correspondence as 

\DeclarePairedDelimiter\floor{\lfloor}{\rfloor}
$j^{**} = \sum_{i=1}^{S(o,j)-1} L(s,i) +  x(j) \floor*{\frac{L(s,S(o,j))}{L(o,S(o,j))}}$
\label{form:sdfsd}

\def\TL{TL$^{*}$}
Finally, our \emph{translation latency with sentence-alignment catch-up} is TL$^{*}(o,j) = t_{f(o,j)} - t_{f(s,j^{**})}$.
This is then averaged for all output words in the document: \TL $= \frac{1}{|o_I|} \sum_{j=1}^{|o_I]} TL^*(o,j)$.
\hidelong{\footnote{For a set of documents $D$, the \TL $= \frac{\sum_{o,I\in D}\sum_{j=1}^{|o_I]} TL^*(o,j)}{\sum_{o,I\in D}|o_I|}$.}}

\subsection{SLT Evaluation}

We use one ASR system for English and nine SLT systems from English into Czech (three different models differing in the data and training parameters), German (2 different systems), French (2 different systems), Spanish and Russian. All the SLT systems are cascades of an ASR, a punctuator, which inserts punctuation and capitalization to unsegmented ASR output, and a neural machine translation (NMT) from the text. 
\hidelong{The systems and their quality measures are in \cref{tab:scores}.}
DE A, ES, and FR B are NMT adapted for spoken translation as in \perscite{Niehues_2018}. The others are basic sentence-level Transformer NMT connected to ASR.
The ASR is a hybrid DNN-HMM by \perscite{janus-iwslt}.

We evaluate the systems on IWSLT tst2015 dataset. 
\hidelong{We downloaded the referential translation from the TED website as \perscite{arivazhagan2019retranslation}, and removed the single words in parentheses because they were not verbatim translations of the speech, but marked sounds such as applause, laughter, or music. 
}

\hidelong{
\begin{table}[t]
\centering
\caption{Quality measure of the English ASR and SLT systems from English into the target language in the left-most column, on IWSLT tst2015. The letters A, B, C denote different variants of SLT systems with the same target. Translation lag (\TL) is in seconds. AcE is average character erasure, NE is normalized erasure.}
\label{tab:scores}
\small
\begin{tabular}{l|rrrrrr}
\bf SLT                            & \bf  BLEU    & \bf  \TL & \bf  AcE & \bf  NE\\
\hline
EN (ASR)       &58.4747  &       & 29.22     & 5.88 \\
\hline
CZ A          & 17.5441 & 2.226 &  24.20 & 7.05 \\
CZ B   &12.2914 & 2.622 &  29.48 & 5.30 \\
CZ C  &18.1505 & 2.933 &  27.90 & 3.93 \\
\hline
DE A                   &15.2678 & 3.506 &  47.32 & 1.39 \\
DE B   &15.9672 & 1.845 &  38.12 & 5.46 \\
\hline
ES                   &21.8516  &5.429 &  43.30 & 1.49 \\
\hline
FR A & 25.8964 & 1.269 &  31.97 & 3.32 \\
FR B                  &20.5367 & 5.425 &  47.92 & 1.46 \\
\hline
RU  &11.6279 & 3.168 &  31.78 & 4.05 \\
\end{tabular}
\end{table}
}

\subsection{Reset Rate}

\begin{figure}
    \centering
    \includegraphics[width=0.7\columnwidth]{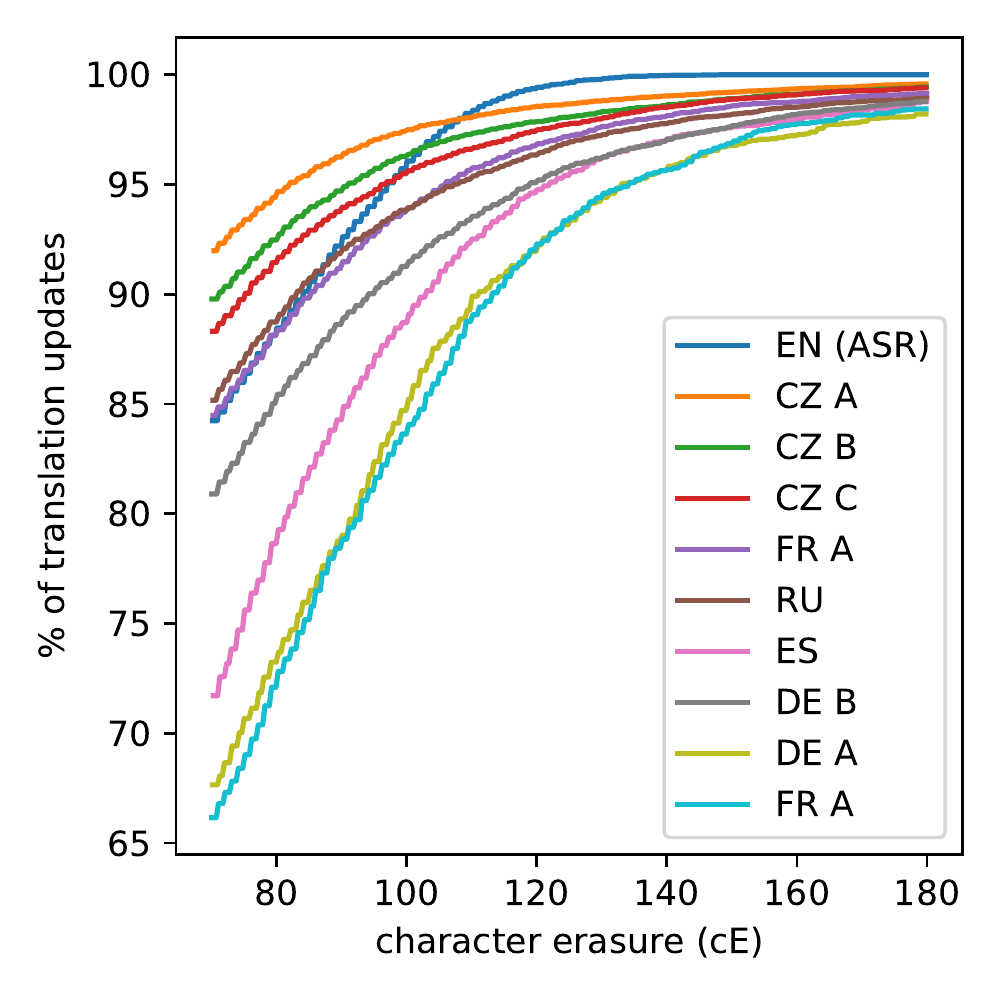}
    \caption{The percentage of translation updates in the validation set with the character erasure less than or equal to the value on the $x$-axis, for all our ASR and SLT systems.%
    \hideXXX{Nesla by osa x nazvat jednoduseji? Nebo aspon napsat: DM: jde to pripsat.} 
    The $x$-axis corresponds with the size of the subtitling window.}
    \label{fig:cdf}
\end{figure}


The average character erasure does not reflect the frequency and size of the individual erasures. Therefore, in \cref{fig:cdf}, we display the cumulative density function of character erasure in the dataset. The vertical axis is the percentage of all translation updates, in which the character erasure was shorter or equal than the value on the horizontal axis. E.g., for the subtitler window with a total size of 140 characters, 99.03 \% of SLT updates of the SLT CZ A fit into this area.
\cref{tab:cdf} displays the same for selected sizes, which fit into 1, 2, and 3 lines of subtitler window of size 70, and also the percentage of updates without any erasure ($x=0$).

The values approximate the expected number of resets. However, the resets are also affected by the blanking time, so the real number of resets may be higher if the speech contains long pauses. The percentage in \cref{fig:cdf} serves as a lower bound.


\hidelong{
\begin{table}[]
\small
    \centering
        \caption{Percentage of character erasures in all translation updates, which are shorter or equal than $x$ characters, for selected values of $x$.}
    \label{tab:cdf}
    \begin{tabular}{l|ccccc}
    SLT & $x = 0$ & $x = 70$ & $x = 140$ & $x = 210$ \\
    \hline
EN (ASR)                           & 20.76 & 84.23 & 99.96 & 100.00 \\
CZ A                               & 41.37 & 91.98 & 99.03 & 99.76 & \\
CZ B                               & 28.61 & 89.78 & 98.63 & 99.77 & \\
CZ C                               & 30.93 & 88.31 & 98.53 & 99.72 & \\
FR A                               & 31.65 & 84.47 & 98.14 & 99.51 & \\
RU                                 & 35.42 & 85.17 & 97.82 & 99.38 & \\
ES                                 & 29.01 & 71.71 & 97.08 & 99.43 & \\
DE B                               & 27.89 & 80.90 & 97.05 & 99.38 & \\
DE A                               & 30.85 & 67.65 & 95.83 & 99.13 & \\
FR A                               & 30.39 & 66.15 & 95.67 & 99.39 & \\
    \end{tabular}
\end{table}
}

\subsection{Subtitling Latency}

\hidelong{
\begin{figure}[t]
    \centering
    \includegraphics[width=0.7\columnwidth]{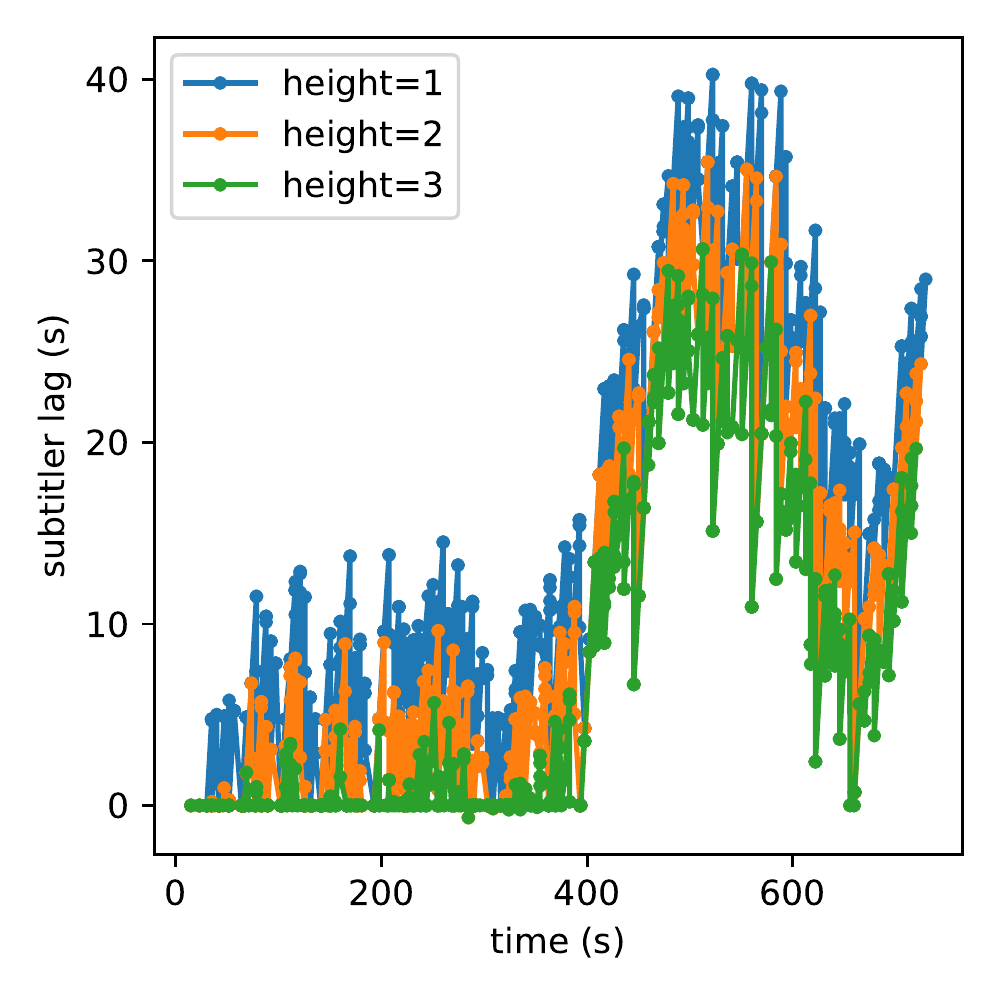}
    \caption{Subtitling latency (y-axis) over time (x-axis) for tst2015.en.talkid1922 translated by CZ A. 
    The subtitling window has the width 70 and height 1, 2 and 3 lines. The minimum reading time is 15 characters per second (one line per 4.7s).} 
    \label{fig:subtitler-lag}
\end{figure}
}

The \emph{subtitling latency} is the difference of the finalization time of a word in subtitler and in the SLT. 
\hidelong{We count it similarly as the translation latency, but the word correspondence is the identity function because the language in SLT and subtitler is the same.}

We computed the latency caused by the subtitler with 1, 2, and 3 lines of width 70 for one talk and SLT systems, see \cref{fig:subtitler-lag}. 
Generally, the bigger the translation window, the lower latency. 

\hideXXX{
The subtitling is useful only when the latency is continuously low. We can observe that in the case of the talk tst2015.en.talkid1922 translated by CZ A (\cref{fig:subtitler-lag}, top), the latency is around zero for 3-line window and below 10 seconds for 2-line window only at the first 400 seconds of the talk. After that, the latency is increasing up to 40 seconds, so the subtitling is desynchronized with the original and probably useless. In the second example (bottom plot), the subtitling is useless from the point where the talk starts.
}

\hideXXX{The long latency may be induced by the low stability of SLT, by the length of incoming segments, by the talking speed, or by other reasons. Our primary motivation in this work is to propose how to estimate the usefulness of the simultaneous subtitling of SLT in a limited area. We leave the analysis of subtitling latency for future work.}

\subsection{User Evaluation}

\begin{table}[]
    \centering
        \caption{Results of user evaluation with three subtitling windows of different heights (h). Quality level 4 is the highest, 1 is the lowest. The right-most column is the percentage of erasures fitting into the subtitling window.}
    \label{tab:rating}
    \small
    \begin{tabular}{r|r@{~}r@{~}r@{~}r|r}
 & \multicolumn{4}{c|}{Percentage of quality levels} \\
height & level=1 & level=2 & level=3 & level=4 & cE $<70\cdot h$\ \\
\midrule
$h = 1$  &  35.27 \% & 28.79 \% & 14.95 \% & 20.99 \% & 88.59 \% \\
$h = 2$  &  11.08 \% & 29.94 \% & 35.73 \% & 23.24 \% & 98.73 \% \\
$h = 3$  &  16.33 \% & 19.90 \% & 33.67 \% & 30.11 \% & 99.64 \% \\
    \end{tabular}
\end{table}

We asked one user to rate the overall fluency and stability of subtitling for 
the first 7-minute part of tst2015.en.talkid1922 translated by CZ A. We presented the user with the subtitles three times, in a window of width 70 and heights 1, 2 and 3. The minimum reading time parameter was 15 characters per second. The user was asked to express his subjective quality assessment by pressing one of five buttons: undecided (0), horrible (1), usable with problems (2), minor flaws, but usable (3), and perfect (4). The user was asked to press them simultaneously with reading subtitles, whenever the assessment changes. The source audio or video was not presented, so this setup is comparable to situations where the user does not understand the source language at all. 
The user is a native speaker of Czech. 


\cref{tab:rating} summarizes the percentage of the assessed duration and the quality levels. The user has not used the level undecided (0). The main problem that the user reported was limited readability due to resets and unstable translations. The flaws in usable parts of subtitling were subtle changes of subtitles which did not distract from reading the new input, or disfluent formulations.

In the right-most column of \cref{tab:rating} we show the percentage of erasures in the part of the evaluated document which fit into the subtitling window. 
We hypothesize that the automatic measure of character erasure may be used to estimate the user assessment of readability.


\section{Conclusion}
\label{conclusion}

We proposed an algorithm for presenting automatic speech translation simultaneously in the limited space of subtitles. The algorithm is independent of the SLT system. It ensures the minimum level of stability and allows simultaneity.
Furthermore, we propose a way of estimating the reader's comfort and overall usability of the SLT with subtitling in limited space, and observe correspondence with user rating. Last but not least, we suggested a catch-up based on sentence-alignment in ASR and SLT to measure the translation latency simply and realistically.



\section*{Acknowledgments}

The research was partially supported by 
the grant CZ.07.1.02/0.0/0.0/16\_023/0000108 (Operational Programme -- Growth Pole of the
Czech Republic), 
H2020-ICT-2018-2-825460 (ELITR) of the EU, 
398120 of the Grant Agency of Charles University,  
and by SVV project number 260 575. 

\small
\bibliography{biblio}
\bibliographystyle{IEEEtran}  


\end{document}